\documentclass[conference,a4paper]{IEEEtran}
\IEEEoverridecommandlockouts

\usepackage[hidelinks]{hyperref}
\usepackage[cmex10]{amsmath}
\usepackage{amssymb,amsfonts}
\usepackage{dblfloatfix}

\usepackage[ruled,vlined]{algorithm2e}
\usepackage{graphicx}
\graphicspath{{Figures/PDF/}{Figures/PNG/}}

\usepackage{booktabs}
\usepackage{siunitx}
\usepackage[numbers,compress]{natbib}
\usepackage{texnames}
\usepackage{bm,bbm}
\usepackage{orcidlink}

\begin{document}

\title{\uppercase{Learning from Noisy Pseudo-labels for All-Weather Land Cover Mapping}
}

\author{
    \IEEEauthorblockN{
        \begin{minipage}{0.33\textwidth}
            \centering
            Wang Liu\orcidlink{0000-0002-0378-7625}\\
            \textit{Hunan University}\\
            410082 Changsha, China\\
            liuwa@hnu.edu.cn
        \end{minipage}%
        \begin{minipage}{0.33\textwidth}
            \centering
            Zhiyu Wang\orcidlink{0009-0004-5113-6132}\\
            \textit{Hunan University}\\
            410082 Changsha, China\\
            wangzhiyu.wzy1@gmail.com
        \end{minipage}%
        \begin{minipage}{0.33\textwidth}
            \centering
            Xin Guo\orcidlink{0000-0003-1448-8978}\\
            \textit{Hunan University}\\
            410082 Changsha, China\\
            flyinggx@hnu.edu.cn
        \end{minipage}%
    }
    \\
    \IEEEauthorblockN{
        \begin{minipage}{0.33\textwidth}
            \centering
            Puhong Duan\orcidlink{0000-0001-5066-4399}\\
            \textit{Hunan University}\\
            410082 Changsha, China\\
            puhong\_duan@hnu.edu.cn
        \end{minipage}%
        \begin{minipage}{0.33\textwidth}
            \centering
            Xudong Kang\orcidlink{0000-0002-3807-2531}\\
            \textit{Hunan University}\\
            410082 Changsha, China\\
            xudong\_kang@163.com
        \end{minipage}%
        \begin{minipage}{0.33\textwidth}
            \centering
            Shutao Li\orcidlink{0000-0002-0585-9848}\\
            \textit{Hunan University}\\
            410082 Changsha, China\\
            shutao\_li@hnu.edu.cn
        \end{minipage}%
    }
    \thanks{This work was supported in part by the National Key Research and Development Program of China under Grant 2021YFA0715203; in part by the National Natural Science Foundation of China under Grant 62201207 and 62371185; in part by the Natural Science Foundation of Hunan Province under Grant 2023JJ40163; in part by the Science and Technology Innovation Program of Hunan Province under Grant 2023RC3124 and  2024RC1030.}
}

\maketitle
\begin{abstract}
Semantic segmentation of SAR images has garnered significant attention in remote sensing due to the immunity of SAR sensors to cloudy weather and light conditions. Nevertheless, SAR imagery lacks detailed information and is plagued by significant speckle noise, rendering the annotation or segmentation of SAR images a formidable task. Recent efforts have resorted to annotating paired optical-SAR images to generate pseudo-labels through the utilization of an optical image segmentation network. However, these pseudo-labels are laden with noise, leading to suboptimal performance in SAR image segmentation. In this study, we introduce a more precise method for generating pseudo-labels by incorporating semi-supervised learning alongside a novel image resolution alignment augmentation. Furthermore, we introduce a symmetric cross-entropy loss to mitigate the impact of noisy pseudo-labels. Additionally, a bag of training and testing tricks is utilized to generate better land-cover mapping results. Our experiments on the GRSS data fusion contest indicate the effectiveness of the proposed method, which achieves first place. The code is available at https://github.com/StuLiu/DFC2025Track1.git.
\end{abstract}


\section{Introduction}
Land-cover mapping is the process of classifying and delineating Earth's surface features,
such as forests, water bodies, and urban areas, using remote sensing technologies.
It plays a vital role in environmental monitoring, sustainable resource management, and city planning
by providing critical data for informed decision-making.
Previous work has almost always used multi-spectral or hyper-spectral data to conduct related research
because these modals can provide rich spatial and spectral information.
However, they cannot work in conditions such as cloudy weather or nighttime due to their reliance on sunlight.
Thus, developing alternative solutions for all-weather land-cover mapping remains an urgent challenge.

Utilizing SAR images to conduct all-weather land-cover mapping has garnered significant attention
in remote sensing due to the immunity of SAR sensors to weather and light conditions \cite{OEM-SAR,SUN1,SUN2}.
To achieve this goal, the 2025 IEEE GRSS Data Fusion Contest fosters the development of
innovative solutions for all-weather land-cover mapping using SAR and optical EO data.
The data consists of multi-modal sub-meter-resolution optical and SAR images with 8-class land-cover pseudo-labels.
These pseudo-labels are derived from optical images based on pre-trained optical image segmentation models.
During the evaluation phase, models will rely exclusively on SAR to ensure they perform well in real-world, all-weather scenarios.
However, the generated pseudo-labels are full of errors, which hinders the effectiveness of the land-cover mapping model.

\begin{figure*}[t!]
	\centering
	\includegraphics[width=0.88\linewidth]{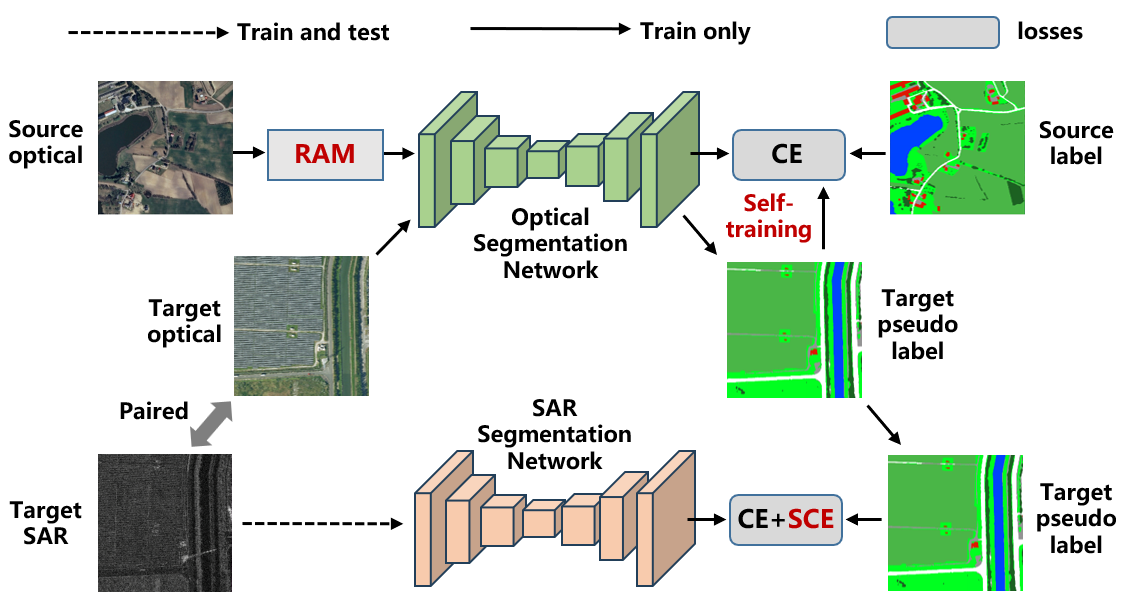}
	\caption{Illustration of the proposed method for the all-weather land-cover mapping task. RAM indicates the resolution alignment augmentation. CE is the cross-entropy loss. SCE represents the symmetric cross-entropy loss. Self-training is one of the most popular domain adaptation techniques.}\label{fig-method}
\end{figure*}

In developing an all-weather load-cover mapping model,
a significant challenge is generating high-quality pseudo-labels for SAR images.
Additionally, training a robust SAR image segmentation model
in condition of a large amount of label noise poses another critical hurdle.
To overcome the above challenges and get better results,
we propose an all-weather land-cover mapping method learning from noisy labels.
This method can be separated into a domain-adaptive optical image segmentation stage
and a SAR image segmentation stage.
Specifically, in the first stage, we employ a domain adaptation paradigm
that can narrow the domain gap between the labeled source data and unlabeled target data to boost performance.
To narrow the distribution discrepancy at the image level,
we propose a resolution alignment augmentation (RAA),
which can randomly downscale the high-resolution images to low-resolution ones.
Furthermore, we introduce the self-training technique to align feature distributions.
In the second stage, we propose a threshold-free pseudo-label selection approach to generate reliable pseudo-labels.
Moreover, to suppress the side-effect of noise labels,
we leverage the symmetric cross entropy (SCE) loss \cite{SCE} to guide network training.
Besides, some commonly utilized tricks in semantic segmentation are used.
Finally, our method gets outstanding performance, achieving first place in the GRSS DFC 2025.

\section{Method}
\subsection{Overview}
In this work, we divide the all-weather land-cover mapping task into two stages.
As shown in Fig. \ref{fig-method}, a domain adaptive semantic segmentation network for optical images
is trained to generate high-quality pseudo-labels in the first stage.
An image-level resolution alignment augmentation is proposed
to align cross-domain features at the image level.
Furthermore, a simple domain adaptation method is introduced
to mitigate the domain gap between the source and target domains at the feature and output levels.

In the second stage, we employ the symmetric cross-entropy loss to alleviate the harmfulness of noisy pseudo-labels. Besides, a set of training tricks is utilized to enhance the robustness of the SAR image semantic segmentation network.

\subsection{Self-training with Resolution Alignment Augmentation}
In this work, we employ the DACS \cite{DACS} as the basic self-training paradigm.
It contains a teacher network $f_{\hat \theta }$ and a siamese student network $f_{\theta }$.
The teacher model aims at generating pseudo-labels,
while the student network is to be trained for optical image segmentation.
Given the source images of OEM \cite{OEM} ${{\cal X}_{\rm S}}{\rm{ = }}\left\{ {{\bf X}_{\rm S}^i} \right\}_{i = 1}^{{N_{\rm S}}}$ with labels ${{\cal Y}_{\rm S}}{\rm{ = }}\left\{ {{\bf Y}_{\rm S}^i} \right\}_{i = 1}^{{N_{\rm S}}}$ and the target images of OEM-SAR \cite{OEM-SAR} ${{\cal X}_{\rm T}}{\rm{ = }}\left\{ {{\bf X}_{\rm T}^i} \right\}_{i = 1}^{{N_{\rm T}}}$ without labels, our method focus on training a well-performed student network for target domain.

In order to mitigate the domain gap at the image level, we propose the resolution alignment augmentation (RAA).
Given an source domain image ${{\bf X}_{\rm S}^i}\in {{\cal R}^{H \times W}}$, we firstly downsample it to smaller-size image:
\begin{equation}
\label{equ_raa_down}
\begin{aligned}
{\bf{X}}_{{\rm{S - D}}}^i = {\cal D}\left( {{\bf{X}}_{\rm{S}}^i,{\mathop{\rm U}\nolimits} \left( {{r_1},{r_2}} \right)} \right)
\end{aligned}
\end{equation}
where $H$ and $W$ indicate the height and the width of the input image ${\bf X}_{\rm S}^i$.
${\bf{X}}_{{\rm{S - D}}}^i$ is the downsampled image.
${\cal D\left( \right)}$ indicates the downsampling function.
${\rm U}\left( {{r_1},{r_2}} \right)$ indicates the random selection with a uniform distribution to get a downsampling ratio.
Sequentially, we upsample the downsampled image to the original size:
\begin{equation}
\label{equ_raa_up}
\begin{aligned}
{\bf{X}}_{{\rm{S - U}}}^i = {\cal U}\left( {{\bf{X}}_{{\rm{S - D}}}^i,H,W} \right)
\end{aligned}
\end{equation}
where ${\bf{X}}_{{\rm{S - U}}}^i\in {{\cal R}^{H \times W}}$ indicates the upsampled image.
${\cal U}()$ is the upsampling function.
In our experiments, the downsampling or upsampling function is set to be bilinear interpolation.
It should be noted that the proposed RAA is
randomly applied during training, and isn't applied to every image.

A standard cross-entropy loss is employed to train the student network:
\begin{equation}
\label{equ_segloss_src_stu}
\begin{aligned}
{{\cal L}_{\rm{S}}} =  - \sum\limits_{i = 1}^B {\sum\limits_{j = 1}^{H \times W} {\sum\limits_{c = 1}^C {{\bf{Y}}_{\rm{S}}^{i,j,c}\log {f_\theta }{{\left( {{{\bf{X}}_{\rm{S}}}|{{\bf{X}}_{{\rm{S - U}}}}} \right)}^{i,j,c}}} } }
\end{aligned}
\end{equation}
where $B$ indicates the batch size within a mini-batch.
$C$ denotes the number of categories.
${\bf{Y}}_{\rm S}$ is the annotation of the source images ${\bf X}_{\rm S}$.
'$|$' represents the or operation.
$f_{ \theta}$ represents the forward process of the student network.
The teacher network is updated by the student network with an exponentially moving average (EMA) algorithm as follows:
\begin{equation}
\label{equ_ema}
\begin{aligned}
{{\hat \theta }_{t + 1}} = \alpha {{\hat \theta }_t} + (1 - \alpha ){\theta _t}
\end{aligned}
\end{equation}
where ${\hat \theta }$ and ${\theta}$ indicate the parameters of the teacher and student networks, respectively.
$\alpha$ is the updating factor for keeping history value and is set to be 0.999.

The teacher network is leveraged to generate pseudo-labels:
\begin{equation}
\label{equ_pseudo}
\begin{aligned}
{{{\bf{\hat Y}}}_{\rm{T}}} = {\rm argmax} \left( {{f_{\hat \theta }}\left( {{{\bf{X}}_{\rm{T}}}} \right)} \right)
\end{aligned}
\end{equation}
Then, we leverage the generated pseudo-labels to train the student network by a weighted cross-entropy loss:
\begin{equation}
\label{equ_segloss_tgt_stu}
\begin{aligned}
{{\cal L}_{\rm{T}}} =  - \sum\limits_{i = 1}^B {\sum\limits_{j = 1}^{H \times W} {\sum\limits_{c = 1}^C {{\lambda ^i}{\bf{\hat Y}}_{\rm{T}}^{i,j,c}\log {f_\theta }{{\left( {{{\bf{X}}_{\rm{T}}}} \right)}^{i,j,c}}} } }
\end{aligned}
\end{equation}
where ${\lambda ^i}$ is the weight factor, which indicates the pseudo-label confidence of the $i$th target image.
It is computed as follows:
\begin{equation}
\label{equ_tgt_weight}
\begin{aligned}
{\lambda ^i} = \frac{1}{{H \cdot W}}\sum\limits_{j = 0}^{H \cdot W} {\left[ {\mathop {\max }\limits_{1 \le c \le C} {f_{\hat \theta }}{{\left( {{{\bf{X}}_{\rm{T}}}} \right)}^{i,j,c}} \ge {\tau _0}} \right]}
\end{aligned}
\end{equation}
where $\left[  \cdot  \right]$ indicates the Iverson bracket. ${\tau}_0$ is the confidence threshold set to be 0.968 following previous work \cite{DACS}.

The total loss function for the student model in the self-training stage is as follows:
\begin{equation}
\label{equ_segloss_stage1}
\begin{aligned}
{{\cal L}_{{\rm{optical}}}} = {\cal L}_{\rm{S}} + {\cal L}_{\rm{T}}
\end{aligned}
\end{equation}

\subsection{Symmetric Cross Entropy Loss}
In order to alleviate the side effects of the noisy pseudo-labels, we introduce the symmetric cross-entropy loss:
\begin{equation}
\label{equ_loss_sar_sce}
\begin{aligned}
{{\cal L}_{{\rm{sce}}}} =  - \sum\limits_{i = 1}^B {\sum\limits_{j = 1}^{H \times W} {\sum\limits_{c = 1}^C {\left( {{f_{{\theta _{{\rm{sar}}}}}}{{\left( {{{\bf{X}}_{{\rm{sar}}}}} \right)}^{i,j,c}}\log \left( {{\bf{\hat Y}}_{\rm{T}}^{i,j,c} + \varepsilon } \right)} \right)} } }
\end{aligned}
\end{equation}
where $f_{{\theta _{{\rm{sar}}}}}$ is the forward function of the segmentation network for SAR images. $\varepsilon$ is a small float value for avoiding log zero error.
A standard cross-entropy loss is applied as the basic objective function:
\begin{equation}
\label{equ_loss_sar_ce}
\begin{aligned}
{{\cal L}_{{\rm{ce}}}} =  - \sum\limits_{i = 1}^B {\sum\limits_{j = 1}^{H \times W} {\sum\limits_{c = 1}^C {\left( {{\bf{\hat Y}}_{\rm{T}}^{i,j,c}\log \left( {{f_{{\theta _{{\rm{sar}}}}}}{{\left( {{{\bf{X}}_{{\rm{sar}}}}} \right)}^{i,j,c}}} \right)} \right)} } }
\end{aligned}
\end{equation}
The final objective function of SAR image segmentation is formulated as follows:
\begin{equation}
\label{equ_loss_sar}
\begin{aligned}
{{\cal L}_{{\rm{sar}}}} = {{\cal L}_{{\rm{ce}}}} + {{\cal L}_{{\rm{sce}}}}
\end{aligned}
\end{equation}

\subsection{A Set of Tricks}
To get more precise segmentation results, a set of tricks is utilized.
Firstly, a larger model size commonly brings better performance.
Secondly, we set the input image size as large as possible under the condition of keeping the batch size to 8 because the model can receive larger context information.
Thirdly, a Lovasz loss is employed to alleviate the class-imbalance problem.
Moreover, a larger number of training iterations usually yields better performance.
Last but not least, an output-level ensemble strategy is employed to boost performance,
which is a simple average of several model predictions.

\begin{table*}[t!]
\centering
\footnotesize
\caption{Results of the ablation study for architectures, backbones, pseudo-labels,
image sizes, the number of iterations, and losses. 'official' means the pseudo-labels provided by previous work. 'ours' indicates the pseudo-labels generated by our method. \\'*' represents the ensembled models.}\label{tab-ablation}
\resizebox{\textwidth}{!}{%
    \begin{tabular}{cccccccc}
        \toprule
        \textbf{configurations} & \textbf{Architectures} & \textbf{Backbones} & \textbf{Pseudo-labels} & \textbf{Image-sizes} & \textbf{Iter-nums} & \textbf{Losses} & \textbf{mIoU (val)}\\
        \midrule
        1  & SegFormer  & MiT-B3 & official & $768 \times 768$ & 40000   & CE & 32.35 \\
        2  & SegFormer  & MiT-B5 & official & $768 \times 768$ & 40000   & CE & 32.72 \\
        3  & SegFormer  & MiT-B3 & ours     & $768 \times 768$ & 40000   & CE & 33.15 \\
        4  & SegFormer  & MiT-B3 & official & $1024 \times 1024$ & 40000 & CE & 32.56 \\
        5  & SegFormer  & MiT-B3 & official & $768 \times 768$ & 160000  & CE & 33.92 \\
        6  & SegFormer  & MiT-B3 & official & $768 \times 768$ & 40000   & CE+Lovasz & 32.64 \\
        7  & SegFormer  & MiT-B5 & ours     & $768 \times 768$ & 160000  & CE+Lovasz & 35.12 \\
        8  & SegFormer  & MiT-B5 & ours     & $768 \times 768$ & 160000  & CE+Lovasz+SCE & 35.31 \\
        9  & SegFormer* & MiT-B5 & ours     & $896 \times 896$ & 160000  & CE+Lovasz+SCE & 35.54 \\
        10 & SegFormer* & MiT-B5   & official+ours  & $896 \times 896$   & 160000 & CE+Lovasz+SCE & 35.70 \\
        11 & UperNet*   & SwinV2-B  & official+ours  & $768 \times 768$   & 160000 & CE+Lovasz+SCE & 35.57 \\
        12 & UperNet*   & ConvNextV2-B  & official+ours & $1024 \times 1024$ & 160000 & CE+Lovasz+SCE & 36.03\\
        \midrule
        13 & Ensemble   & - & -          & - & - & - & \textbf{36.64} \\
        \bottomrule
    \end{tabular}%
}
\end{table*}

\section{Experiment Setting}
\subsection{Datasets}
We utilized the OpenEarthMap \cite{OEM} (OEM) and OpenEarthMap-SAR \cite{OEM-SAR} (OEM-SAR) datasets as the source and target domains to train the semantic segmentation networks in stage one. These datasets contain the identical category space: background, bareland, rangeland, developed space, road, tree, water, agricultural land, and building. Specifically, OEM contains 3000 training and 500 validating optical images with detailed annotations. OEM-SAR contains 4333 paired optical and SAR images with pseudo-labels for training, 210 images for validating, and 490 images for testing. Note that there is a large domain gap between the images of OEM and the ones of OEM-SAR, especially the image resolution discrepancy.

\subsection{Experiment Configs}
We utilize the SegFormer \cite{SegFormer} and UperNet \cite{UperNet} as the basic segmentation network architectures. The MixVisionTransformer (MiT) \cite{SegFormer}, Swin-TransformerV2 \cite{SwinV2}, and ConvNextV2 \cite{ConvNextV2} are employed as the backbones. The batch size is set to 8. The images fitted to the networks are set to $768 \times 768$, $896 \times 896$, or $1024 \times 1024$, which is determined by the model size. The random resized crop, flip, rotate, and color jitter are utilized to constitute our basic data augmentation. All the experiments are conducted with four Nvidia RTX-4090-24GB GPUs. The performance will be evaluated using the mean intersection over union (mIoU) metric.

\section{Results and Analysis}

\subsection{ablation study}
\noindent{\bf Key components.}
In this section, we conduct experiments to analyze the effectiveness of the key components in our method, including model architectures (Architectures), backbones, pseudo-labels, input image size (Image-sizes),
the number of training iterations (Iter-nums), and loss functions (Losses).
The ablation experiment results are shown in the Table. \ref{tab-ablation}.
Configuration 1 is the baseline of our experiments, and it achieves $32.35\%$ mIoU score in the validating set.
A slight improvement is gained after replacing the backbone MiT-B3 with MiT-B5
confirming larger model size can improve the performance.
In configuration 3, we change the official pseudo-labels to the ones generated by our method.
We can see that a significant improvement is gained,
which indicates the effectiveness of our domain adaptation approach.
Surprisingly, similar enhancements are observed while improving the input image size,
increasing the number of training iterations, or introducing the class-balanced Lovasz loss.
For configuration 8, we add the SCE loss, and a satisfactory improvement is achieved, suggesting that suppressing high-confidence examples is helpful when training with the noisy pseudo-labels.
Finally, we train the SegFormers and UperNets with diverse backbones, and we ensemble their predictions. The ensemble prediction gets a 36.64\% mIoU score in the validating set and a 40.08\% mIoU score on the testing set.

\subsection{Comparing with Previous Work}
We compare our method with the official benchmark, and the results are listed in the Table. \ref{tab-sota}.
It reveals that our method outperforms the official benchmark 3.75\% mIoU score in the testing set.
The prediction ensemble is quite useful.
\begin{table}[t!]
\centering
\caption{Results on the testing set when comparing with previous work.}\label{tab-sota}
\begin{tabular*}{\linewidth}{@{\extracolsep{\fill}}cccc@{}}
    \toprule
    \textbf{Methods} & \textbf{Architectures} & \textbf{Backbones} &  \textbf{mIoU (test)} \\
    \midrule
    Official benchmark \cite{OEM-SAR}  & UNet      & -     & 35.13 \\
    Official benchmark \cite{OEM-SAR}  & SegFormer & -     & 35.77 \\
    Official benchmark \cite{OEM-SAR}  & VMamba    & -     & 34.74 \\
    \midrule
    Ours     & SegFormer    &  MiT-B5           & 39.52 \\
    Ours     & UperNet      &  SwinV2-B         & 40.58 \\
    Ours     & UperNet      &  ConvNextV2-B     & 40.18 \\
    Ours (Ensemble)     & -     & -             & \textbf{41.08} \\
    \bottomrule
\end{tabular*}%
\end{table}

\section{Conclusion}
In this work, an effective two-stage land-cover mapping solution in all-weather conditions is proposed. Specifically, our image resolution alignment strategy aligns the distribution between the annotated and unlabeled data. Therefore, the model is easily adapted to the target domain, and more accurate pseudo-labels are generated. Furthermore, the noise pixels are suppressed, and the semantic segmentation networks are converged to a better point after employing symmetric cross-entropy loss. Finally, by incorporating the training and testing tricks, the land-cover mapping results are boosted. In the future, the vision foundation model for SAR images will serve as the image encoder, assisting high-quality land-cover mapping.

\small
\bibliographystyle{IEEEtranN}
\bibliography{references}

\end{document}